\documentclass{article} \PassOptionsToPackage{numbers,sort&compress}{natbib} \usepackage[preprint]{neurips_2026} \usepackage[utf8]{inputenc} \usepackage[T1]{fontenc} \usepackage{hyperref} \usepackage{url} \usepackage{booktabs} \usepackage{amsmath} \usepackage{amsfonts} \usepackage{nicefrac} \usepackage{microtype} \usepackage{xcolor} \usepackage{graphicx} \usepackage{multirow} \usepackage{color,soul}
\title{Can Time Series Foundation Models Serve as Surrogates for Power Systems Dynamic Simulation?}
\author{Anonymous Authors}
\author{%
Sandy Miguel \quad Bruno Paes Leao\thanks{Corresponding author} \quad Cheng Feng \\
Siemens Data \& AI Research\\
\texttt{\{sandy.miguel,bruno.leao,cheng.feng\}@siemens.com}\\
}

\begin{document}
\maketitle

\begin{abstract} High-fidelity dynamic simulation is essential for scientific and engineering applications, but its computational cost limits large-scale scenario screening. We investigate the use of pretrained time-series foundation models (TSFMs) as computationally efficient surrogates to address this limitation. In-context learning and post-training strategies, including retrieval-augmented reference sample selection and physics-informed regularization, are considered in the investigation. Results are benchmarked against a traditional AI model representative of the state-of-the-art in the domain. Experiments using an open-source TSFM and data from a simulated 9-bus power system indicate promising results. Retrieval augmentation improves in-context learning performance, and fine-tuning leads to superior results compared to the traditional AI model, especially when including physics-informed regularization. The fine-tuned TSFMs achieve lower errors and improved computational costs compared to the traditional AI model, and a 563$\times$ speedup compared to the original physics-based simulation. \end{abstract}

\section{Introduction}
Recent time-series foundation model (TSFM) literature such as \citet{ansari2025chronos} shows that large-scale pretraining can yield general-purpose predictors that adapt across diverse forecasting tasks. This progress raises a broader question for scientific and engineering applications: can TSFMs act as fast surrogates for computationally intensive simulations while preserving the fidelity required for decision-making? This potential is particularly attractive for power systems planning, where engineers routinely evaluate post-disturbance behavior across thousands of hypothetical operating conditions, events, and future grid configurations. At the same time, modern grids are presenting more complexity due to high penetration of inverter-based controls that lead to faster transient dynamics, and growing load variability from data centers and electric vehicle charging. High-fidelity simulators such as ANDES \cite{cui2024andes} and PSS/E \cite{psse_software} remain the standard, but their runtimes make comprehensive screening computationally expensive or even infeasible. Existing machine-learning surrogates \cite{wang2025multi, shahbaz2026causal} offer substantial speedups, yet they are typically tailored to specific systems and operating conditions. 

Recent TSFMs such as Chronos-2 \cite{ansari2025chronos} have demonstrated strong forecasting capabilities across diverse domains. However, their ability to adequately simulate the physically-constrained dynamics of multi-bus power systems has not yet been verified. This paper makes four key contributions: (1) providing evidence of the applicability of time-series foundation models as fast surrogates for post-disturbance power-system trajectory simulation; (2) assessing how strategies for in-context learning and post-training can enhance such capability; (3) introducing physics-informed regularization in TSFM post-training; (4) benchmarking the proposed strategies against a physics-based simulator and a state-of-the-art traditional AI solution.

\section{Methodology}
\textbf{Problem formulation.} A power system is naturally represented as a graph whose buses are coupled through the transmission network. A disturbance at one location therefore propagates across the system and affects voltage, angle, and frequency trajectories at multiple buses. Given historical measurements $\mathbf{Y}_{1:t}$ and known past and future disturbance information represented by covariates $\mathbf{C}_{1:T}$, the objective is to simulate the future system trajectory over the horizon $T-t$:
\begin{equation}
\label{eq:problem_formulation}
\hat{\mathbf{Y}}_{t+1:T}=F_{\phi}(\mathbf{Y}_{1:t},\mathbf{C}_{1:T}),
\end{equation}
where $F_{\phi}(\cdot)$ is the data-driven simulator and $\hat{\mathbf{Y}}_{t+1:T}$ contains the predicted voltage magnitude ($V$), voltage angle ($\theta$), frequency ($f$), and rate of change of frequency ($\mathrm{ROCOF}$) trajectories at all buses.

In this work, we focus specifically on load disturbances. The future covariates $\mathbf{C}_{t+1:T}$ describe the scenario to be simulated through the active- and reactive-power load profiles ($P$ and $Q$) at the applicable buses. Thus, the model receives as inputs not only the pre-disturbance system trajectory but also information defining the future disturbance. This formulation allows real-world events such as load pickup/rejection, industrial or motor-load changes, and reactive-power stress conditions to be represented. More generally, the framework is not restricted to load events: other types of events such as generator outages, line trips, or faults could be represented by appropriate definition of the covariates.

\textbf{In-Context Learning for Dynamic Trajectories:} Employing a TSFM for the defined simulation task is achieved by means of its forecasting capability, i.e. eq. \eqref{eq:problem_formulation} is interpreted as a forecasting problem. Unlike conventional forecasting models that require task-specific training, TSFMs can adapt to new tasks through \emph{in-context learning}. In this case, a context window $W \in \phi$ is defined, corresponding to the concatenation in time of multiple reference samples $S$, i.e. $W=concat([S^1, S^2, ..., S^K])$, where $concat(.)$ corresponds to the time concatenation operation and $K$ is the number of reference samples employed to create $W$. Each reference sample $S$ corresponds to a limited time window of multivariate time series obtained from the physics-based simulator. Given a simulation query, this context window is extended by concatenation of the known part of the test sample $Y_{1}$ and provided to the TSFM. For the specific problem considered here, each reference sample consists of a disturbance scenario containing both pre-disturbance observations and the post-disturbance response. This is how the TSFM can be informed of the patterns defining the behavior of the system of interest and perform the task defined in eq. \eqref{eq:problem_formulation} in a zero/few-shot fashion - zero-shot from the model training perspective and few-shot from the in-context learning perspective.

\textbf{Strategies for employing the TSFM.} Besides the base methodology described above, we investigate two groups of strategies to improve TSFMs performance: (i) enhanced in-context learning by means of reference sample selection strategies, and (ii) fine-tuning, using training data generated by the physics-based simulator.

\textbf{1. Enhanced In-Context Learning.} Two strategies are employed for the selection of reference samples for the context window, assuming the availability of a set $\Sigma$, such that $S^i \in \Sigma, \forall i$ and $|\Sigma|>K$: (i) uniform random sampling, referred to below as \textbf{random}; and (ii) retrieval-based selection, which picks the nearest neighbors of the test sample using cosine similarity to the embeddings of covariate variables ($C$) as the distance metric. This is referred to below as \textbf{RAG}.

\textbf{2. Fine-Tuning.} The scope of fine-tuning considered here targets adaptation of the TSFM for the specific application, and not the creation of a foundation model for power systems simulation. Therefore, the focus in this case is not on zero/few shot generalization, but rather in investigating whether using a pre-trained TSFM as starting point can lead to improved performance compared to training a traditional AI model from scratch using the same training data. We consider two fine-tuning variants: (i) \emph{standard fine-tuning} based on the same loss function employed during the TSFM pre-training, referred to below as \textbf{standard FT}. (ii) \emph{physics-informed regularization} where we augment the forecasting objective with a physics-based regularization term derived from power-grid physics, referred to below as \textbf{physics FT}. The hypothesis is that this inductive bias can improve the performance of the post-trained model. Power-balance residuals have previously been used to train neural power-flow surrogates by penalizing violations of Kirchhoff's laws \cite{donon2020graph}. The final loss function is:
\begin{equation}
\label{eq:total_loss}
\mathcal{L}_{\mathrm{total}}
=
\mathcal{L}_{\mathrm{forecast}}
+
\lambda \mathcal{L}_{\mathrm{phys}}.
\end{equation}
Predicted voltage magnitudes $\hat{V}_b(\tau)$ and voltage angles
$\hat{\theta}_b(\tau)$ are used to compute the corresponding active and reactive
power injections, $P_b^{\mathrm{calc}}(\tau)$ and $Q_b^{\mathrm{calc}}(\tau)$,
through the AC power-flow equations. The physics regularizer penalizes
power-balance residuals at passive and load buses:
\begin{align}
\label{eq:physics_loss}
\mathcal{L}_{\mathrm{phys}}
={}&
\frac{1}{|\mathcal{T}|\,|\mathcal{B}_{\mathrm{p}}|}
\sum_{\tau\in\mathcal{T}}
\sum_{b\in\mathcal{B}_{\mathrm{p}}}
\left(
\left|P_b^{\mathrm{calc}}(\tau)\right|
+
\left|Q_b^{\mathrm{calc}}(\tau)\right|
\right)
\nonumber\\[-1mm]
&+
\frac{1}{|\mathcal{T}|\,|\mathcal{B}_{\mathrm{l}}|}
\sum_{\tau\in\mathcal{T}}
\sum_{b\in\mathcal{B}_{\mathrm{l}}}
\left(
\left|P_b^{\mathrm{calc}}(\tau)+P_b^{\mathrm{sched}}(\tau)\right|
+
\left|Q_b^{\mathrm{calc}}(\tau)+Q_b^{\mathrm{sched}}(\tau)\right|
\right).
\end{align}
Here, $\mathcal{B}_{\mathrm{p}}$ and $\mathcal{B}_{\mathrm{l}}$ denote the sets
of passive and load buses, respectively, while
$P_b^{\mathrm{sched}}(\tau)$ and $Q_b^{\mathrm{sched}}(\tau)$ denote the
scheduled active and reactive load consumption at bus $b$ and timestep $\tau$.
$\mathcal{L}_{\mathrm{phys}}$ corresponds to a self-supervised term that acts as a soft constraint, encouraging physically consistent
forecasts while preserving the flexibility required to represent transient
power-system dynamics.

\section{Experimental Evaluation}
\subsection{Experimental Setup}
We evaluated post-disturbance trajectory simulation on the IEEE 9-bus system \cite{anderson2003power}, containing three generators, nine buses, and six load points. A total of 1,800 samples were generated using ANDES across six load-related disturbance modes, including normal load pickup/rejection, motor/industrial load pickup, and reactive-power voltage-stress events. 1,622 samples were used for reference or training/validation and the remaining 178 were held out for testing. Each sample contains 436 time steps at 60~Hz ($7.26$~s): 180 pre-event steps (3~s) and 256 disturbance/post-event steps (4.26~s). Targets comprise $(V,\theta,f,\mathrm{ROCOF})$ at nine buses (36 channels), while covariates comprise $(P,Q)$ at six load buses (12 channels). Normal and stressed operating conditions are equally represented, where the latter include under/over-frequency or under/over-voltage violations relative to the stated operating ranges \cite{ANSI_C84_1_2020}.
\subsection{Evaluation Methodology}

The Chronos-2 open-source TSFM \cite{ansari2025chronos} was selected for the investigation due to its demonstrated performance in benchmarking with other similar models and especially due to its ability to take into consideration future values of the covariates for forecasting. Another desirable characteristic is that Chronos-2 can provide stochastic estimates based on selected quantiles. As a traditional AI baseline, we trained a conditional spatiotemporal graph neural network (STGNN) combining graph convolutional layers with GRU cells, based on recent literature \cite{wang2025multi,shahbaz2026causal}. It uses 64-dimensional hidden states and two GNN layers, and is trained for 100 epochs with batch size 32, learning rate $10^{-3}$, and MSE loss.

\textbf{In-Context Learning:} For each query, a context window with $K=12$ reference samples was selected from the available reference scenarios and concatenated with the pre-disturbance observations of the test sample. We investigated the two reference-selection strategies previously described - random selection and RAG-based retrieval - using the 178 held-out samples for testing.

\textbf{Fine-Tuning:} Chronos-2 was fine-tuned via LoRA (rank $r=32$, scaling $\alpha=64$) targeting six transformer layers. Training proceeded for 10,000 steps with effective batch size 192, learning rate $5 \times 10^{-5}$ with cosine annealing, and gradient clipping at norm 5.0. In the \textbf{standard FT} variant, the model was optimized using the native multi-quantile pinball loss \cite{ansari2025chronos}. In the \textbf{physics FT} variant, the physics-based regularization term in Eq.~\eqref{eq:physics_loss} was added after a 1,000-step warmup with a 2,000-step ramp. We evaluated $\lambda \in \{5\times10^{-4},\,1\times10^{-3},\,2\times10^{-3}\}$ and selected $\lambda=5\times10^{-4}$ based on validation performance. Fine-tuned models were trained on 1,431 samples, validated on 191, and tested on the 178 held-out samples. We also trained the same Chronos-2 architecture using the same recipe but initialized with random weights to isolate the benefit of pretraining.

\textbf{Metrics:} We report MAE on point prediction and CRPS for probabilistic estimates, averaged across the relevant buses and timesteps. Inference time was measured as an average of 10 runs. AI models were tested using NVIDIA H200 GPUs, while ANDES was tested on a server with Intel Xeon Gold 6248 12-core CPUs.

\subsection{Results}
Table~\ref{tab:performance} summarizes the results. Within the zero/few-shot setting, RAG reduced CRPS for all targets and MAE for three of the four targets (except ROCOF) compared to random context selection, indicating that scenario-relevant reference trajectories generally improve in-context learning. However, the trained STGNN achieved better performance than zero/few-shot Chronos-2 on all point estimates.

\begin{table}[!htb]
\centering
\caption{Performance comparison. All methods are evaluated on the same 178 held-out samples.}
\label{tab:performance}

\small
\setlength{\tabcolsep}{2.5pt}
\renewcommand{\arraystretch}{1.08}

\begin{tabular*}{\textwidth}{
    @{\extracolsep{\fill}}
    l
    *{8}{c}
    @{}
}
\toprule
\multirow{2}{*}{\textbf{Method}}
& \multicolumn{2}{c}{\textbf{$V$ (pu)}}
& \multicolumn{2}{c}{\textbf{$\theta$ (deg)}}
& \multicolumn{2}{c}{\textbf{$f$ (Hz)}}
& \multicolumn{2}{c}{\textbf{ROCOF (Hz/s)}} \\
\cmidrule(lr){2-3}
\cmidrule(lr){4-5}
\cmidrule(lr){6-7}
\cmidrule(lr){8-9}
& MAE & CRPS
& MAE & CRPS
& MAE & CRPS
& MAE & CRPS \\
\midrule
Chronos-2 Random
& 0.008863 & 0.00488
& 1.1237   & 0.638
& 0.2163   & 0.140
& 0.09723  & 0.0724 \\

Chronos-2 RAG
& 0.003780 & 0.00251
& 0.4200   & 0.243
& 0.1012   & 0.0647
& 0.09923  & 0.0553 \\
\midrule

STGNN Traditional AI
& 0.001094 & --
& 0.2223   & --
& 0.06900  & --
& 0.01633  & -- \\
\midrule

Chronos-2 Standard FT
& 0.000437 & 0.00140
& \textbf{0.06183} & 0.235
& 0.01091  & 0.039
& 0.007901 & 0.030 \\

Chronos-2 Physics FT
& \textbf{0.000402} & \textbf{0.00120}
& 0.06283 & \textbf{0.198}
& \textbf{0.01013} & \textbf{0.037}
& \textbf{0.007865} & \textbf{0.028} \\
\bottomrule
\end{tabular*}
\end{table}

\begin{figure}[!htb]
    \centering
    \includegraphics[width=1\textwidth]{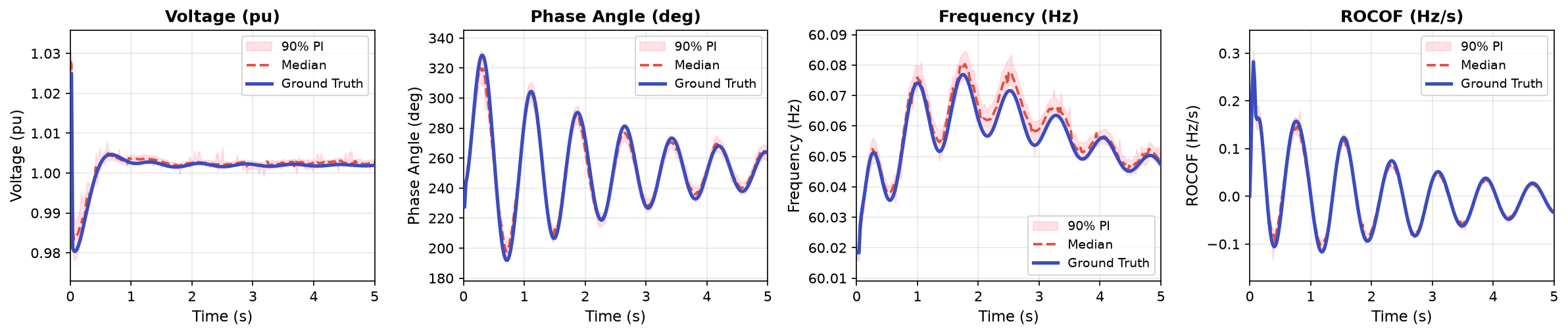}
    \caption{Representative trajectories from physics-informed fine-tuned Chronos-2 under a stressed power-system condition.}
    \label{fig:physics_ft_example}
\end{figure}

Fine-tuning provided a substantial improvement. Both fine-tuned Chronos-2 variants outperformed the STGNN on the held-out set. This, combined with the fact that the randomly initialized control experiment failed to converge on all attempts, supports the hypothesis that TSFM pretraining provides a strong starting point for a grid-specific dynamic surrogate compared to learning the complete temporal representation from scratch. Physics FT was superior to Standard FT in every reported metric except $\theta$ MAE, where the difference is small, suggesting that the power-balance inductive bias improves both point accuracy and probabilistic calibration. Figure~\ref{fig:physics_ft_example} presents an illustrative example of a test sample with ground truth, point and stochastic estimates.

From a computational perspective, Chronos-2 inference time (35.5ms) was \textbf{563$\times$ faster} than the 20s ANDES runtime. It was also considerably faster than the STGNN model (187.1ms) due to the autoregressive nature of the traditional AI model.

\section{Conclusion and Future Work}
This work provides evidence that pretrained TSFMs can serve as accurate, efficient, and uncertainty-aware surrogates for power system dynamic simulation. Tests with Chronos-2 indicated that retrieval-based in-context learning improves zero/few-shot performance without parameter updates, while fine-tuning yields substantially better performance than the zero/few-shot variants and also achieves superior results compared to the traditional AI baseline. Physics-informed fine-tuning yields the best overall performance, indicating that grid physics can be incorporated effectively as a post-training inductive bias while retaining fast inference. Future work will include experiments towards larger systems, cross-topology and cross-disturbance generalization, and power-system-specialized foundation models. A further direction is trying other forms of post-training learned inductive bias such as the enforcement of hard-constraints (e.g. \cite{utkarsh2026hardconstraints}).

\begin{ack}
The authors would like to thank Danielle Maddix Robinson for the valuable support and advice, and the anonymous reviewers for the insightful comments which contributed to improve the quality of this work.
\end{ack}

\bibliographystyle{unsrtnat}
\bibliography{refs}

\end{document}